\documentclass[lettersize,journal]{IEEEtran}
\usepackage{amsmath,amsfonts}
\usepackage{algorithmic}
\usepackage{algorithm}
\usepackage{array}
\usepackage[caption=false,font=scriptsize,labelfont=sf]{subfig}
\usepackage{textcomp}
\usepackage{stfloats}
\usepackage{url}
\usepackage{verbatim}
\usepackage{graphicx}
\usepackage{cite}
\usepackage{bm}
\usepackage{accents}
\usepackage{array}
\usepackage{multicol}
\usepackage{multirow}
\usepackage{eqparbox}
\usepackage{enumerate}

\newcolumntype{M}[1]{>{\centering\arraybackslash}m{#1}}
\newcolumntype{P}[1]{>{\centering\arraybackslash}p{#1}}

\newcommand\munderbar[1]{%
  \underaccent{\bar}{#1}}

\begin{document}

\title{\fontsize{20.5}{24.5}\selectfont Deep Reinforcement Learning for Optimal Power Flow with Renewables Using Spatial-Temporal Graph Information}

\author{Jinhao Li,
Ruichang Zhang,
Hao Wang, ~\IEEEmembership{Member~IEEE,}
~\\
Zhi Liu, ~\IEEEmembership{Member~IEEE,}
Hongyang Lai,
Yanru Zhang, ~\IEEEmembership{Member~IEEE}
\thanks{Manuscript received April 19, 2021; revised August 16, 2021.}
\thanks{J. Li, H. Lai are with the School of Mechanical and Electrical Engineering, University of Electronic Science and Technology of China, Chengdu, China, 611731 (e-mail:jinhaoli@std.uestc.edu.cn, laihongyang0314@outlook.com).}
\thanks{R. Zhang, Y. Zhang are with the School of Computer Science and Engineering, University of Electronic Science and Technology of China, Chengdu, China, 611731. Y. Zhang is also affiliated with Shenzhen Institute of Advanced Study, University of Electronic Science and Technology of China. (e-mail:yanruzhang@uestc.edu.cn, zhangruichang2@gmail.com). Y. Zhang is the corresponding author.}
\thanks{H. Wang is with the Department of Data Science and Artificial Intelligence, Faculty of Information Technology, Monash University, Melbourne,VIC 3800, Australia (e-mail: hao.wang2@monash.edu).}
\thanks{Z. Liu is with Department of Computer and Network Engineering, University of Electro-Communications, Tokyo, Japan (e-mail:liu@ieee.org).}
\vspace{-2em}
}

\markboth{Journal of \LaTeX\ Class Files,~Vol.~14, No.~8, August~2021}%
{Shell \MakeLowercase{\textit{et al.}}: A Sample Article Using IEEEtran.cls for IEEE Journals}


\maketitle
\begin{abstract}
Renewable energy resources (RERs) have been increasingly integrated into large-scale power systems. Considering uncertainties and voltage fluctuation issues introduced by RERs, in this paper, we propose a deep reinforcement learning (DRL)-based strategy \emph{leveraging spatial-temporal (ST) graphical information of power systems}, to dynamically search for the optimal power flow (OPF), of power systems with a high uptake of RERs. We formulate the OPF problem as a multi-objective optimization problem considering generation cost, voltage fluctuation, and transmission loss, and employ deep deterministic policy gradient (DDPG) to learn an optimal strategy for OPF. Moreover, given that the nodes in power systems are self-correlated and interrelated in temporal and spatial views, we develop a \emph{multi-grained attention-based spatial-temporal graph convolution network (MG-ASTGCN) for extracting ST graphical correlations and features}, aiming to provide graphical knowledge of power systems for its sequential DDPG in our DRL algorithm to more effectively solve OPF. We validate our algorithm on modified IEEE 33, 69, and 118-bus radial distribution systems and demonstrate that our algorithm outperforms other benchmark algorithms. Our experimental results also reveal that our MG-ASTGCN can significantly accelerate DDPG's training process and performance in solving OPF. The proposed DRL-based strategy also improves power systems’ robustness in the presence of node faults, especially for large-scale power systems.
\end{abstract}


\section{Introduction} \label{section_introduction}
\IEEEPARstart{T}{here} has been an exponential growth of distributed renewable energy resources (RERs) in smart grids for mitigating global climate change and providing affordable electricity to customers. From the year 2007 to 2020, the global installed capacity of solar photovoltaic (PV) has increased from $8$ Gigawatts (GW) to $760$ GW, and the wind power capacity has also risen from $94$ GW to $743$ GW \cite{RERs-report}. Although the adoption of RERs in conventional power systems offers various benefits, such as decarbonizing the electricity market and reducing the energy supply costs, the integration of RERs into power systems, at the same time, poses significant challenges due to their intermittent nature \cite{Reddy2017}. 

One of the major challenges is that the RERs lead to consecutive and fast changes of the optimal power flow (OPF) in power systems \cite{IMPRAM2020100539}. Solving the OPF problem is mostly formulated to minimize the cost of power generation while satisfying power systems' operating constraints \cite{opf}. Various external factors, such as changes in solar irradiation and wind velocity, cause stochastic and non-dispatchable generation of RERs, inevitably leading to continuous changes of power flow. Moreover, since the number of newly installed RER systems cannot be accurately predicted, several technical issues can also occur, such as voltage fluctuations and harmonic distortions, threatening power systems' stability and can cause potential economic losses \cite{Shafiullah2010}. These challenges motivate our work to take uncertainty factors into consideration and to optimize power flow efficiently for smart grids with a high penetration of RERs.

The existing studies on solving OPF can be briefly categorized into three classes---traditional, model-based, and learning-based algorithms, which are less likely to address challenges caused by RERs. 1) traditional methods, such as Gauss-Seidel algorithm \cite{literature_opf_gauss-seidel}, Newton-Raphson algorithm \cite{literature_opf-newton-raphson}, interior point methods \cite{literature_opf_interior}, and etc., have proved their excellent performance in power systems supplied by only fuel energy resources. However, these methods are extremely difficult to converge because of the uncertainties brought by RERs when solving the OPF problem \cite{literature_opf_traditional-drawback}; 2) Model-based approaches, including stochastic and robust optimization \cite{literature_opf_model-based-stochastic-optim-1,literature_opf_model-based-stochastic-optim-2,literature_opf_model-based-robust-optim-1,literature_opf_model-based-robust-optim-2}, have been introduced to solve the OPF problem leveraging metaheuristic algorithms, e.g., artificial bee colony algorithm \cite{literature_opf_model-based-abc}, harris hawk optimization (HHO) \cite{hho}, and grey wolf optimization (GWO) \cite{gwo}. Nonetheless, model-based methods can be less effective when it comes to large-scale power systems, as most of model-based methods are highly dependent on accurate knowledge of given power systems and sensitive to initialization values  \cite{Siano2012,Nusair2020,Nusair2020-2}. Moreover, model-based algorithms suffer from a heavy computational burden and can trap into local optimum \cite{hho}; 3) Learning-based methods, such as machine-learning and deep-learning algorithms, 
Deep-learning methods require a large amount of historical data to train an accurate deep-learning model \cite{Lei2021,deep-learning-method,8909795}, becoming a barrier to the adoption of deep-learning algorithms. In particular, it is inherently challenging for deep-learning methods to react quickly to dynamic changes of optimal operating point.

To overcome the aforementioned drawbacks, this paper introduces deep reinforcement learning (DRL) to solve the OPF problem. DRL is well suitable to capture the dynamic features in RERs-rich power systems, and thus powerful in solving the OPF problem \cite{Cao2021}. However, learning a stable and well-performed DRL-based strategy is time-consuming due to its slow convergence in complex systems, such as OPF of power systems \cite{rl_book}. How to extract effective information from complex power systems becomes an essential to accelerate the learning of DRL algorithms. Due to the strong coupling in OPF problems, the nodes (e.g., buses) in power systems are self-correlated and interrelated in temporal and spatial views, containing a prior knowledge of power systems that can play a significant role in assisting the DRL model to solve OPF more effectively. Existing studies did not capture such effective features, e.g., the spatial-temporal (ST) information, about power flows. Therefore, we develop a multi-grained attention-based spatial-temporal convolution network (MG-ASTGCN) to extract ST correlations through attention mechanism and ST features through graph convolution, providing effective information for the DRL algorithm.

The main contributions of our work are summarized as follows.
\begin{itemize}
    \item \emph{Solving OPF by DRL}: We propose a DRL-based strategy using DDPG leveraging ST graphical information to solve OPF with a high renewable penetration in power systems. Our method can dynamically dispatch power flow, search for the optimal operating point, and quickly respond to uncertainties brought by RERs. Our DRL algorithm is demonstrated to outperform benchmark algorithms, such as harris hawk optimization (HHO) and grey wolf optimization (GWO), through simulations. Moreover, our DRL-based strategy improves power systems' stability in the presence of node faults, especially for large-scale power systems.
    \item \emph{Extraction of Spatial and Temporal Information}:
    We develop the multi-grained attention-based spatial-temporal graph convolution network (MG-ASTGCN) to fully extract ST information in dynamic power flow, where the attention mechanism and graph convolution are adopted for mining ST correlations and features, respectively. Additionally, given that the power flow exhibits periodic patterns in different time scales, we construct mutli-grained power flow time-series to capture multi-scale ST information. MG-ASTGCN can provide global prior knowledge of power systems for the DRL algorithm to accelerate its convergence in solving OPF.
\end{itemize}

The key insights drawn from our work are summarized as follows.
\begin{itemize}
    \item \emph{DDPG converges faster with the assistance of MG-ASTGCN}: Our experimental results show that the addition of MG-ASTGCN in DDPG can significantly improve DDPG's convergence speed, compared to other correlation extraction methods, which demonstrates effectiveness of the developed MG-ASTGCN in capturing prior ST graphical information in power systems.
    \item \emph{Impacts of the Spatial-temporal Attention Mechanism in MG-ASTGCN}: The spatial attention mechanism in MG-ASTGCN aims to capture correlation strengths among nodes in power systems, while temporal attention exploits temporal dependencies on each node' features. Our experiments reveals that node pairs with more generator access have stronger spatial correlations. In the temporal view, each node's features are highly self-correlated in the past $10$ time intervals.
    \item \emph{Power systems tends to work in the sub-optimal operation point if controlling voltage fluctuation is overemphasized.} The DRL-based strategy encodes all operation constraints of OPF into the reward functions as feedback from power systems. It is found that if the reward function for controlling voltage fluctuation is overemphasized, the performance of DRL-based strategy degenerates simultaneously, which results in sub-optimal operation of power systems.
\end{itemize}

The remainder of this paper is organized as follows. In Section \ref{sec_system-model}, the system model is presented. The DRL-based strategy, including MG-ASTGCN and DDPG, are introduced in Section \ref{sec_methodology}. The experimental setup and results are presented in Section \ref{sec_experiments}. Section \ref{sec_conclusions} concludes this paper.
\begin{figure}[!t]
    \centering
    \vspace*{-3mm}
    \includegraphics[width=3.2in,height=2.6in]{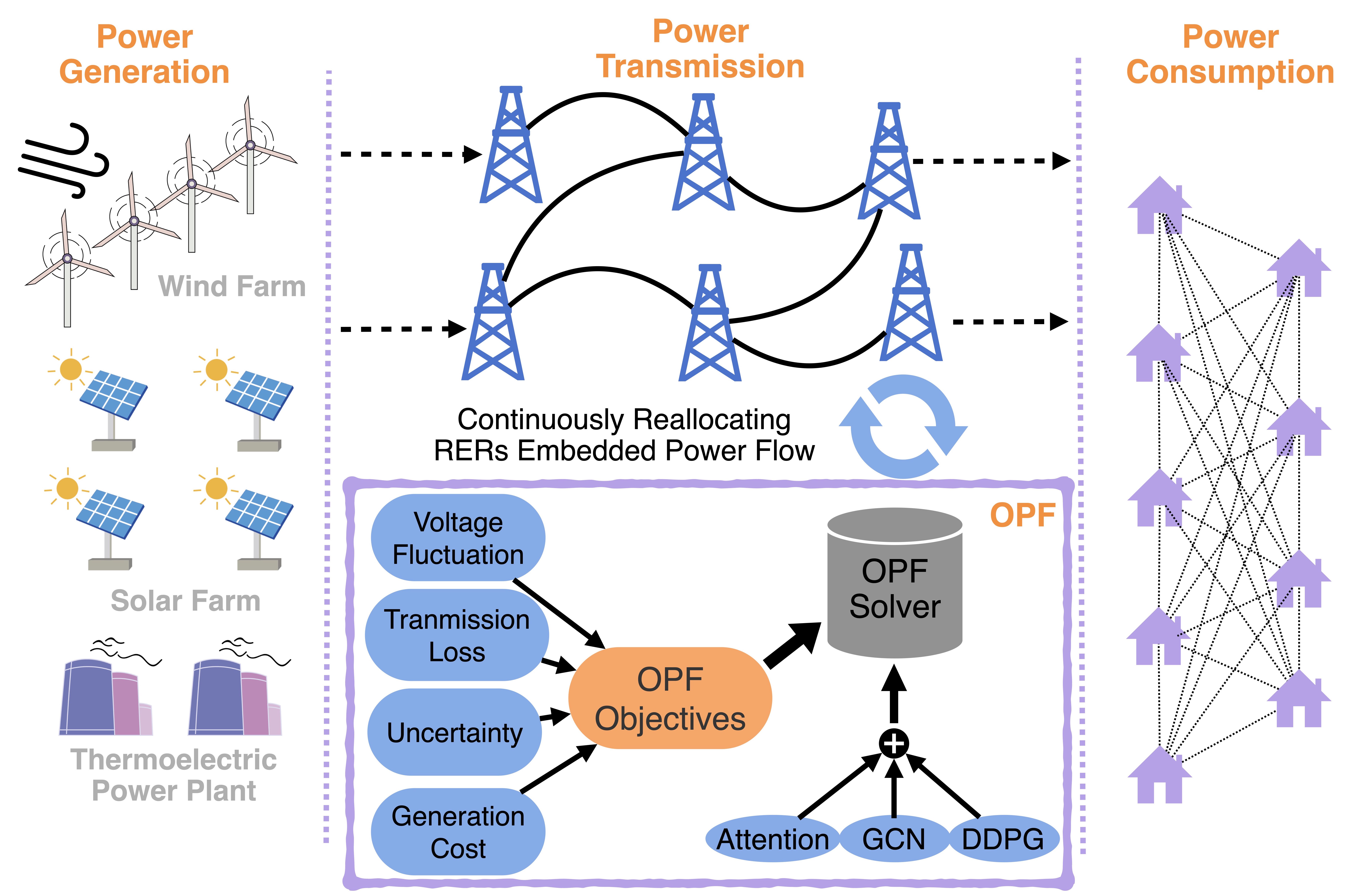}
    \vspace*{-3mm}
    \caption{The system model and the presented work.}
    \label{fig:total_framework}
    \vspace*{-5mm}
\end{figure}

\section{System Model}
\label{sec_system-model}
We consider a distributed power distribution network whose power generation is supported by fuel energy, wind power, and solar PV power. Wind and solar PV power are regarded as the most representative RERs with the largest installed capacities \cite{REN21}. The high uptake of RERs introduces uncertainties and leads to the voltage fluctuation making the OPF problem more challenging. We formulate the OPF problem into a multi-objective OPF (MO-OPF) optimization problem, in which both uncertain factors and voltage fluctuation control are considered. An overview of the presented work is illustrated in Fig. \ref{fig:total_framework}.

\subsection{Power Generation Cost}
\label{subsec_power-gen-cost}
The OPF problem usually aims to minimize the power generation cost while satisfying power systems' operating constraints. For fuel energy resources, considering the valve-point effect modeled as a sinusoidal function \cite{Bouchekara2016}, the generation cost of the $i$th thermoelectric generator can be formulated as
\begin{equation}
\label{eq_thermal-gen-cost}
    C_{\text{t},i} = a_i\left(P_{\text{t},i}\right)^2+b_iP_{\text{t},i}+c_i+|d_i\sin\left(e_i\left(\munderbar{P}_{\text{t},i}-P_{\text{t},i}\right)\right)|,
\end{equation}
where $P_{\text{t},i}$ and $\munderbar{P}_{\text{t},i}$ represent the actual and minimal power outputs, respectively, and $a_i,\cdots,e_i$ are constant coefficients.

Due to the intermittent nature of RERs, the cost related to RERs can be divided into two parts: 1) the direct power generation cost \cite{Biswas2017} and 2) the mismatch cost between scheduled and available power generation \cite{Panda2015}. The direct power generation costs for wind and solar PV power are presented, respectively as
\begin{align}
\label{eq_wind-gen-cost}
C_{\text{w},j} = f_jP_{\text{w},j},\\
\label{eq_solar-gen-cost}
C_{\text{s},k} = g_kP_{\text{s},k},
\end{align}
where $f_j$ and $g_k$ are constant coefficients. $P_{\text{w},j}$ and $P_{\text{s},k}$ represent the scheduled power outputs of the $j$th wind turbine and $k$th solar PV power generator, respectively.

The power mismatch occurs when wind and solar PV power generation are lower or higher than their schedules, resulting in power overestimation or underestimation from uncertain sources, respectively. The variability of RERs can be modeled by an associated probability density function. The power generation cost for power overestimation, namely its reserve cost, can be defined for wind and solar PV power, respectively as
\begin{align}
\label{eq_wind-reserve-cost}
C^\text{r}_{\text{w},j} &=\mathbb{E}\left[h^{\text{r}}_{\text{w},j}\left(P_{\text{w},j}-P^\text{a}_{\text{w},j}\right)\right]\\
&=h^{\text{r}}_{\text{w},j} \int_{\munderbar{P}_{\text{w},j}}^{P_{\text{w},j}} \left(P_{\text{w},j}-p_{\text{w},j}\right) f_\text{w}\left(p_{\text{w},j}\right)d\left(p_{\text{w},j}\right) \nonumber ,\\
\label{eq_solar-reserve-cost}
C^{\text{r}}_{\text{s},k} &= \mathbb{E}\left[h^{\text{r}}_{\text{s},k} \left(P_{\text{s},k}-P^{\text{a}}_{\text{s},k}\right)\right]\\
&=h^{\text{r}}_{\text{s},k} \int_{\munderbar{P}_{\text{s},k}}^{P_{\text{s},k}} \left(P_{\text{s},k}-p_{\text{s},k}\right)f_\text{s}\left(p_{\text{s},k}\right) d\left(p_{\text{s},k}\right)\nonumber,
\end{align}
where $h^{\text{r}}_{\text{w},j}$ and $h^{\text{r}}_{\text{s},k}$ represent constant coefficients. $P^\text{a}_{\text{w},j}$ and $P^{\text{a}}_{\text{s},k}$ represent the available power outputs of the $j$th wind turbine and $k$th solar PV power generator, respectively.

On the contrary, under power underestimation circumstance, if there is no mechanism to reduce power generation from thermoelecrtic generators, the redundant power generation will be curtailed. We define penalty costs for wind and solar PV power, respectively as
\begin{align}
\label{eq_wind-penalty-cost}
C^\text{p}_{\text{w},j} &= \mathbb{E}\left[h^\text{p}_{\text{w},j}\left(P^\text{a}_{\text{w},j}-P_{\text{w},j}\right)\right],\\
\label{eq_solar-penalty-cost}
C^\text{p}_{\text{s},k} &= \mathbb{E}\left[h^\text{p}_{\text{s},k}\left(P^\text{a}_{\text{s},k}-P_{\text{s},k}\right)\right],
\end{align}
where $h^{\text{p}}_{\text{w},j}$ and $h^{\text{p}}_{\text{s},k}$ are defined as coefficients.

The power generation cost, considering thermoelectric generators, wind turbines, and solar PV power generator, is defined as
\begin{equation}
\begin{aligned}
C=\sum_{i=1}^{N_\text{t}}C_{\text{t},i}&+\sum_{j=1}^{N_\text{w}} \left(C_{\text{w},j}+C^\text{r}_{\text{w},j}+C^\text{p}_{\text{w},j}\right) \\
&+\sum_{k=1}^{N_\text{s}}\left(C_{\text{s},k}+C^\text{r}_{\text{s},k}+C^\text{p}_{\text{s},k}\right).
\end{aligned}
\end{equation}

\subsection{Voltage Fluctuation Control}
\label{subsec_voltage-fluctuation}
Voltage fluctuation occurs in power systems especially in the presence of RERs, which greatly degrades the performance of electronic equipment and poses potential security risks on consumers. To mitigate voltage fluctuations, we consider voltage control with metric $F$ defined as
\begin{equation}
\label{eq_voltage_f}
    F = \sum_{j=1}^{N} \left| V_j^{t_0}-\frac{1}{T_{\text{td}}}\sum_{t=1}^{T_{\text{td}}}V_j^{t_0-t} \right|,
\end{equation}
to describe nodes' voltage stability. In Eq. (\ref{eq_voltage_f}), $N$ represents the number of buses in the given power system, including all power generators, and $V_j^{t_0}$ is the $j$th bus voltage at $t_0$.

\subsection{Power Loss}
\label{subsec_constraints}
Transmitting power to consumers inevitably leads to power losses in power systems, and the power losses can be formulated as
\begin{equation}
    L= \sum_{b=1}^{N_\text{b}} G_{i,j}\left[V_i^2+V_j^2-2V_iV_j\cos\left(\delta_{ij}\right)\right],
\end{equation}
where $N_\text{b}$ is the total number of branches in the power system, $\delta_{ij}=\delta_i-\delta_j$ represents the voltage angle difference between the $i$th and $j$th buses, and $G_{i,j}$ is the transfer conductance of the $b$th branch connecting the $i$th and $j$th buses.

\subsection{MO-OPF Formulation}
To minimize power generation cost, mitigate voltage fluctuation, and reduce power loss, we consider them as three objectives of the MO-OPF optimization problem defined in Eq. (\ref{eq_first-objective})-(\ref{eq_third-objective}).
\begin{gather}
\label{eq_first-objective}
\min \hspace{0.25em} C, \\
\label{eq_second-objective}
\min \hspace{0.25em} F,\\
\label{eq_third-objective}
\min \hspace{0.25em} L.
\end{gather}

The above objectives are subject to physical constraints that ensure the safe operation of power systems, which are formulated as
\begin{align}
\label{eq_constraints-1}
P_{gi}-P_{li}&=V_i \sum_{b=1}^{N_\text{b}} V_j\left(G_{i,j}\cos \delta_{ij}+B_{i,j}\sin \delta_{ij}\right),\\ 
\label{eq_constraints-2}
Q_{gi}-Q_{li}&=V_i \sum_{b=1}^{N_\text{b}} V_j\left(G_{i,j}\sin \delta_{ij}+B_{i,j}\cos \delta_{ij}\right),
\end{align}
\begin{align}
\label{eq_constraints-3}
\munderbar{P}_{i} &\leq P_{i} \leq \bar{P}_{i},  &&i=1,\cdots,N,\\
\label{eq_constraints-4}
\munderbar{Q}_{i} &\leq Q_{i} \leq \bar{Q}_{i}, &&i=1,\cdots,N,\\
\label{eq_constraints-5}
\munderbar{V}_{i} &\leq |\mathbb{V}_{i}| \leq \bar{V}_{i},  &&i=1,\cdots,N,\\
\label{eq_constraints-6}
&|\mathbb{S}_b| \leq \bar{S}_b, &&b=1,\cdots,N_{b},
\end{align}
where $\bar{P}_i$ and $\bar{Q}_i$ are the maximal active and reactive power of the $i$th bus, respectively, and $\bar{S}_b$ represents the maximal apparent power on the $b$th branch. Note that both $\mathbb{V}_{i}$ and $\mathbb{S}_b$ are complex numbers of voltage and apparent power, respectively. Note that Eq. (\ref{eq_constraints-1}) and (\ref{eq_constraints-2}) define the power balance constraint, in which both generated active and reactive power must be equal to power consumption and losses \cite{10.5555/559961}.

\section{Methodology}
\label{sec_methodology}
In this section, we propose a DRL-based algorithm to solve the complicated MO-OPF problem described in Eq. (\ref{eq_first-objective})-(\ref{eq_constraints-6}), which is reformulated into a Markov decision process (MDP). DDPG is then adopted to solve the derived MDP for its state-of-the-art performance among various DRL algorithms. Moreover, the MG-ASTGCN is introduced to fully extract ST information in power systems, which assists the learning in the sequential DDPG for better performance.

\subsection{Spatial-Temporal Correlations Extraction via MG-ASTGCN}
\label{subsec_astgcn}
\subsubsection{Preliminaries of MG-ASTGCN}
\begin{figure}[!t]
    \centering
    \includegraphics[width=2.7in,height=2.5in]{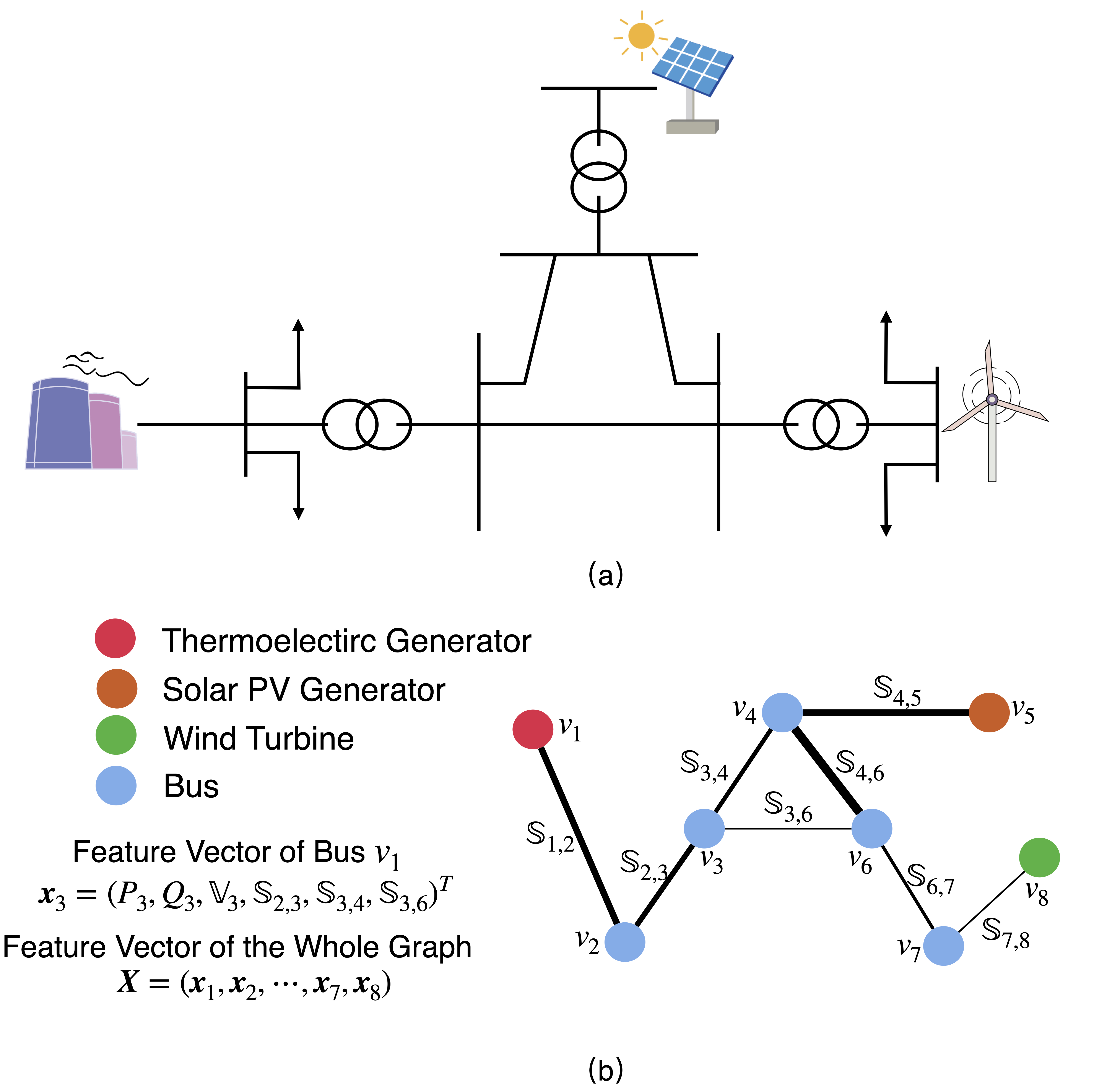}
    \vspace*{-4mm}
    \caption{(a) An actual simplified power system. (b) The transformed graph. Note that the thickness of edges on (b) indicates different values of power flow.}
    \label{fig:system-to-graph}
    \vspace*{-7mm}
\end{figure}

Power system can be modeled as an undirected graph $\mathcal{G}=(\mathcal{V},\mathcal{E},\bm{A})$, as illustrated in Fig. \ref{fig:system-to-graph}. Each node $v_i$ generates a feature vector $\mathbf{x}_i$ for gathering local stationary information in each time interval, as formulated in Eq. (\ref{eq_node-feature-vector}). The aggregate form of feature vector of $\mathcal{G}$ is shown in Eq. (\ref{eq_graph-feature-vector}).
\begin{align}
\label{eq_node-feature-vector}
\mathbf{x}_i &= \left(P_{i},Q_{i},\mathbb{V}_i,\mathbb{S}_{i,j_1},\mathbb{S}_{i,j_2},\cdots\right)^T,\\
\label{eq_graph-feature-vector}
\mathbf{X} &= \left(\mathbf{x}_1,\mathbf{x}_2,\cdots,\mathbf{x}_N\right).
\end{align}

\begin{figure}[!t]
    \centering
    \includegraphics[width=3.4in,height=1.67in]{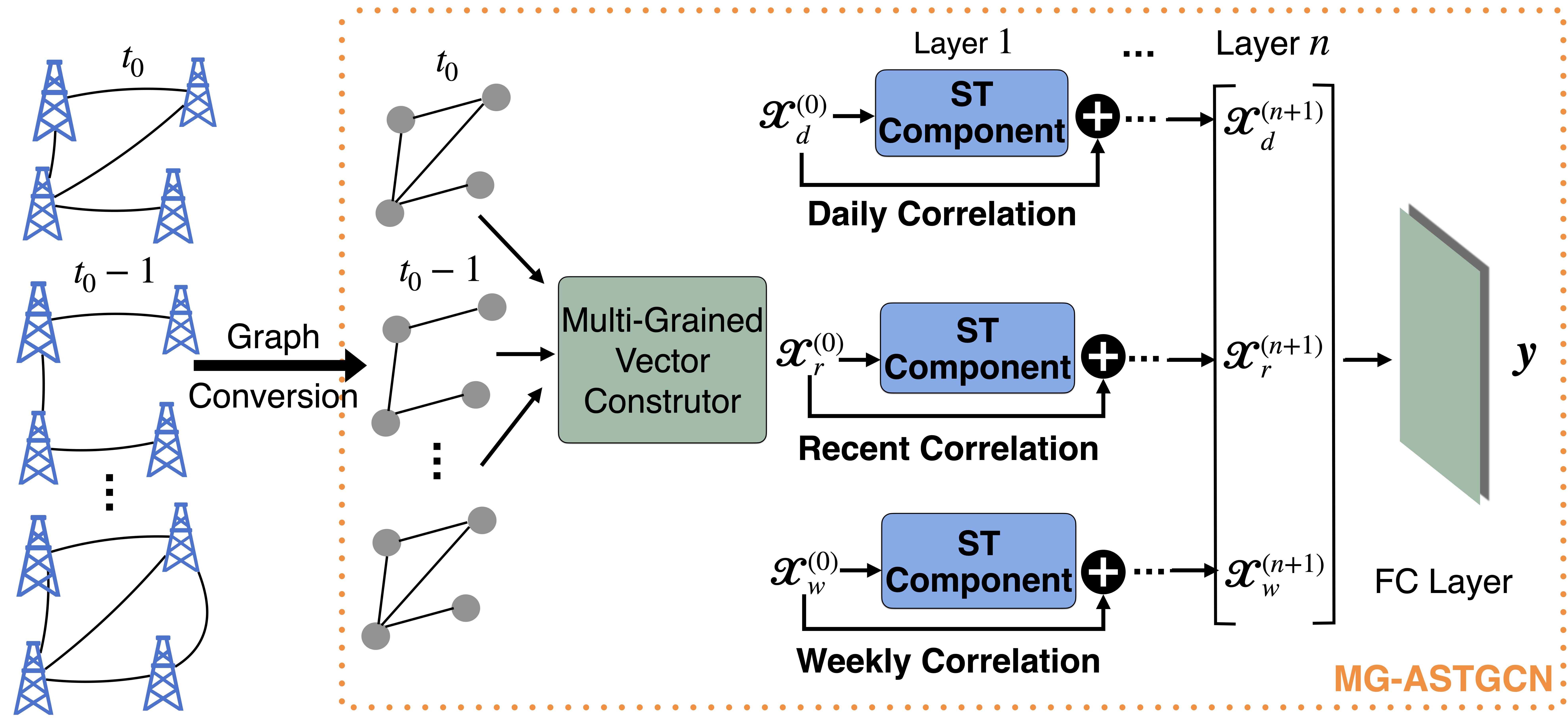}
    \vspace*{-4.5mm}
    \caption{The framework of the proposed MG-ASTGCN.}
    \label{fig:mgastgcn_framework}
    \vspace*{-5mm}
\end{figure}

Considering periodic patterns in power flow \cite{REN21}, e.g., daily and weekly patterns, a multi-grained vector constructor is developed to better capture temporal correlations of a series of $\mathcal{G}$, as shown in Fig. (\ref{fig:mgastgcn_framework}), which divided the $\mathcal{G}$ into recent, daily, and weekly segments as
\begin{align}
\label{eq_recent-segment}
\bm{\mathcal{X}}_\text{r} &= \left\{\mathbf{X}_{t_0-T_\text{r}+1},\cdots,\mathbf{X}_{t_0-1},\mathbf{X}_{t_0}\right\},\\
\label{eq_daily-segment}
\bm{\mathcal{X}}_\text{d} &= \left\{\mathbf{X}_{t_0-T_\text{d}\times n_{\text{d}}},\cdots,\mathbf{X}_{t_0-n_{\text{d}}},\mathbf{X}_{t_0}\right\},\\
\label{eq_weekly-segment}
\bm{\mathcal{X}}_\text{w}&= \left\{\mathbf{X}_{t_0-7\times T_\text{w}\times n_{\text{d}}},\cdots,\mathbf{X}_{t_0-7\times n_{\text{d}}},\mathbf{X}_{t_0}\right\},
\end{align}
where $T_\text{s}$, $T_\text{d}$, $T_\text{w}$ indicate the length of recent, daily, and weekly segments, respectively, and $n_\text{d}$ represents the frequency of adjusting power flow per day.

The framework of the proposed MG-ASTGCN is illustrated in Fig. \ref{fig:mgastgcn_framework}, including graph transformation operation, multi-grained vector constructor, and ASTGCN. ASTGCN is introduced to take multi-grained segments defined in Eq. (\ref{eq_recent-segment})-(\ref{eq_weekly-segment}) as inputs, in which each segment passes through several ST components for ST information extraction. The structure of one ST component is illustrated in Fig. \ref{fig:st_component}.

\subsubsection{Spatial-Temporal Attention Mechanism}
\begin{figure}[!b]
    \centering
    \vspace*{-6mm}
    \includegraphics[width=3.4in,height=1.2in]{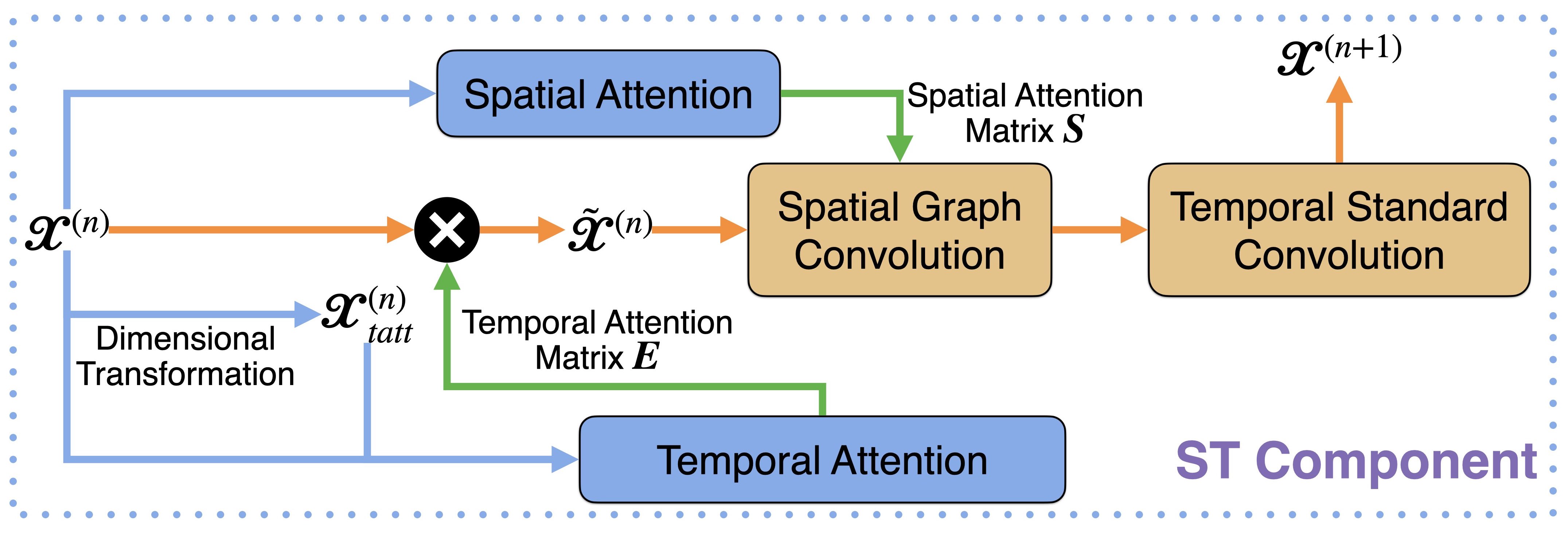}
    \vspace*{-5mm}
    \caption{The structure of one ST component. $\bm{\mathcal{X}}^{n}$ is the output of the previous ST component while $\bm{\mathcal{X}}^{n+1}$ represents the input for the next ST component.}
    \label{fig:st_component}
\end{figure}

The ST attention mechanism is conducted before graph convolution, as shown in Fig. \ref{fig:st_component}. The idea is to pay more attention to valuable graphical information in both spatial and temporal perspectives on $\mathcal{G}$ for the sequential convolution operations, which can be considered as graph preprocessing.

\textbf{Spatial Attention}: Mutual influence between each node and its neighboring nodes varies dynamically due to changes of power flows. Hence, an attention mechanism in spatial dimension is developed to capture the dynamic correlations among nodes \cite{9062547}, which can be formulated as
\begin{gather}
\mathbf{S} = \mathbf{V}_\text{s}\odot \sigma\left[\left(\bm{\mathcal{X}}^{n-1}  \mathbf{W}_\text{t}\right)^T\mathbf{W}_{\text{qk}}\left(\mathbf{W}_\text{f}\bm{\mathcal{X}}^{n-1}\right)^T+\mathbf{b}_\text{s}\right],\\
\label{eq_spatial-att-matrix}
s_{i,j}' = \frac{\exp(s_{i,j})}{\sum_{j=1}^N\exp(s_{i,j})},
\end{gather}
where $\mathbf{V}_\text{s}$, $\mathbf{W}_\text{t}$, $\mathbf{W}_{\text{qk}}$, $\mathbf{W}_\text{f}$, and $\mathbf{b}_\text{s}$ are all learnable parameters. Note that, the sigmoid function $\sigma(\cdot)$ is employed as the activation function. $\mathbf{S}$ represents the spatial attention matrix, whose element $s_{i,j}$, named attention weight, semantically describes the correlation strength between the $i$th and $j$th nodes. Besides, normalized via Softmax operation in Eq. (\ref{eq_spatial-att-matrix}) \cite{GoodBengCour16}, $\mathbf{S}'$ can be adopted for graph convolution to adjust connection weights among nodes.

\textbf{Temporal Attention}: Similar to the above spatial attention, the temporal attention mechanism \cite{attention} can be formulated to track temporal correlations of each node's state as
\begin{gather}
\mathbf{E} = \mathbf{V}_\text{e}\odot \sigma\left[\left(\mathbf{U}_\text{n} \bm{\mathcal{X}}^{n-1}_{\text{tatt}}\right)^T\mathbf{U}_{\text{qk}}\left(\mathbf{U}_\text{f}\bm{\mathcal{X}}^{n-1}\right)+\mathbf{b}_\text{e}\right],\\
\label{eq_temporal-att-matrix}
e_{i,j}'=\frac{\exp(e_{i,j})}{\sum_{j=1}^{T^{n-1}}\exp(e_{i,j})},
\end{gather}
where $\mathbf{V}_\text{e}$, $\mathbf{U}_\text{n}$, $\mathbf{U}_{\text{qk}}$, $\mathbf{U}_\text{f}$, and $\mathbf{b}_\text{e}$ are learnable parameters, and the temporal attention's input $\bm{\mathcal{X}}^{n-1}_{\text{tatt}}$ is the transposed form of $\bm{\mathcal{X}}^{n-1}$ for the convenience of matrix multiplication. The element $e_{i,j}'$ represents the normalized strength of temporal dependency between two graph feature vectors $\mathbf{X}_{t_0-i}$ and $\mathbf{X}_{t_0-j}$. The obtained $\mathbf{E}'$ is used for adding temporal correlations to the ST component's input $\bm{\mathcal{X}}^{n}$, as shown in Fig. \ref{fig:st_component}.

\subsubsection{Spatial-Temporal Convolution}\label{subsubsec_st-conv}
The ST convolution consists of spatial graph convolution and temporal standard convolution, aiming to extract ST features, and reduce dimensions of inputs to be applicable for our DRL algorithm.

\textbf{Spatial Graph Convolution}: Graph convolution is defined as a convolution operation implemented by using linear operators diagonalizing in the Fourier domain to replace the classical convolution operator \cite{gcn}, which can be expressed as
\begin{equation}
\label{eq_graph-conv}
    \begin{aligned}
    \text{ReLU}(g_\theta *_G x) &= \text{ReLU}\left[g_\theta \left(\bm{L}\right) x\right],\\
    &= \text{ReLU}\left[\bm{U}^T\left(\bm{U}x\odot\bm{U}g_\theta\right)\right],
    \end{aligned}
\end{equation}
where $\bm{L}$ is $\mathcal{G}$'s Laplacian matrix, the graph convolution operator is denoted by $*_G$, $g_\theta$ is convolution filter, and the rectified linear unit (ReLU) is adopted as the activation function. Graph convolution is normally processed via eigenvalue decomposition, in which $\bm{U}$ is the result of decomposition. However, such decomposition is computationally expensive. In practice, Chebyshev polynomials are used for efficiently approximating the solution of eigenvalue decomposition \cite{simonovsky2017dynamic}, with the addition of previous obtained spatial attention matrix $s'$ in Eq. (\ref{eq_spatial-att-matrix}), which can be rewritten as
\begin{equation}
\label{eq_graph-conv-rewrite}
    \text{ReLU}(g_\theta *_G x) \approx \sum_{k=0}^{K-1} \theta_k\left[T_k(\tilde{\bm{L}})\odot \bm{S}'\right]x,
\end{equation}
where $\tilde{\bm{L}}$ is the normalized Laplacian matrix, $\theta_k$ is the coefficient of Chebyshev polynomials, and $T_k$ is the $k$th order Chebyshev polynomial.
\begin{figure*}[!t]
    \centering
    \includegraphics[width=7in,height=3.3in]{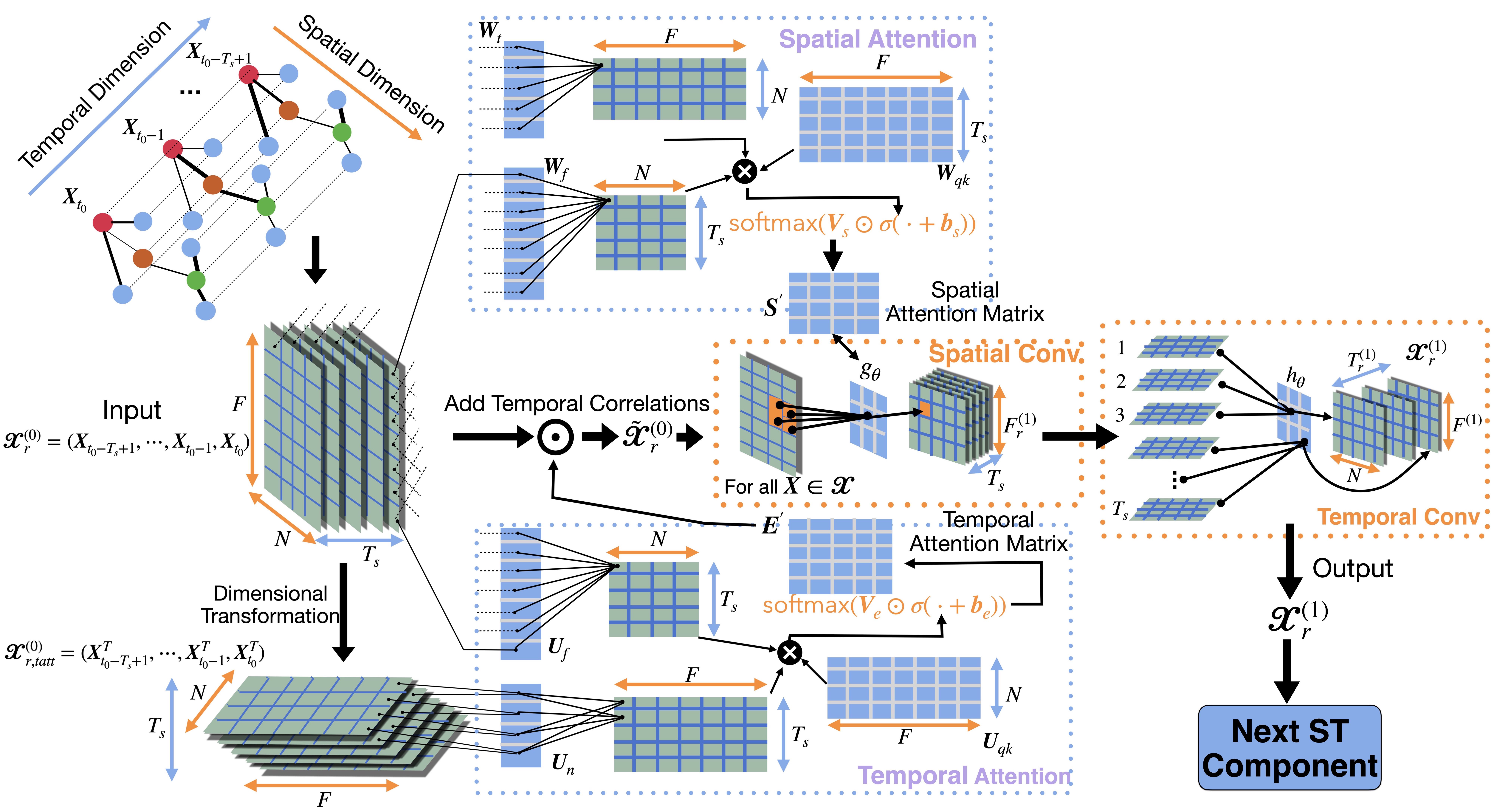}
    \vspace*{-3mm}
    \caption{The detailed process of one recent segment passing through one whole ST component.}
    \label{fig:process_segment}
    \vspace*{-7mm}
\end{figure*}

\textbf{Temporal Convolution and Feature Compression}:
According to the procedure in Fig. \ref{fig:st_component}, the standard convolution operation is conducted in the temporal dimension, taking the result of spatial graph convolution as input, which can be formulated as
\begin{equation}
\label{eq_temporal-conv}
    \bm{\mathcal{X}}^{n+1} = \text{ReLU}\left\{h_\theta * \left[\text{ReLU}\left(g_\theta *_G \tilde{\bm{\mathcal{X}}}^{n}\right)\right]\right\},
\end{equation}
where $h_\theta$ is the standard convolution filter.

To apply the results of MG-ASTGCN as prior graphical knowledge for our DRL algorithm, the outputs in different time scales are fused and then fed into a fully-connected neural network layer for compression, as shown in Fig. \ref{fig:mgastgcn_framework}, which can be presented as
\begin{equation}
\label{eq_output-mgastgcn}
    \bm{y} = \text{ReLU}\left\{\textbf{Dense}\left[\textbf{concat}\left(\bm{\mathcal{X}}_\text{r}^{n+1},\bm{\mathcal{X}}_\text{d}^{n+1},\bm{\mathcal{X}}_\text{w}^{n+1}\right)\right]\right\},
\end{equation}
where \textbf{concat} represents the concatenating operation for the above three outputs, and $\textbf{Dense}$ is referred as a fully-connected layer for feature compression.

To sum up, ST graphical information can be fully exploited by MG-ASTGCN. Firstly, spatial attention figures out node pairs' correlation strength while temporal attention focuses on mining self-correlations of each node's features in the temporal view. Then, ST convolution is introduced to extract valuable features hidden in power systems based on information provided by the preceding ST attention. The detailed process of one segment passing through one whole ST component is illustrated in Fig. \ref{fig:process_segment}.

\subsection{Solving the MO-OPF Problem via DRL}
\subsubsection{MDP Modeling}
Solving MO-OPF can be considered as a consecutive decision-making process. We model the developed MO-OPF problem as a dynamic MDP \cite{rl_book}, consisting of four parts: $\left(\mathcal{S},\mathcal{A},\mathcal{P},\mathcal{R}\right)$.

\textbf{State} $\mathcal{S}$: For the $i$th node in $\mathcal{G}$, its state $s_i^t$ is the feature vector $\mathbf{x}_i^t$ proposed in Eq. (\ref{eq_node-feature-vector}).

\textbf{Action} $\mathcal{A}$: For each power generator, only its active power $P_i^t$ and voltage $V_i^t$ can be manipulated. Thus, the action space of the $i$th node can be expressed as $a_i^t=\{P_i^t,\mathbb{V}_i^t\}$.

\textbf{Probability} $\mathcal{P}$: $\mathcal{P}$ is the probability of a transition to the next state $s_i^{t+1}$ from the current state $s_i^t$ taking selected action $a_i^t$. In DRL, an action strategy is learned to deal with different states, denoted by $\pi:\mathcal{S}\rightarrow \mathcal{P}^\pi(\mathcal{A})$, which maps states to a probability distribution over actions. Note that $\mathcal{P}^\pi$ is different from $\mathcal{P}$, in which the probabilistic strategy $\mathcal{P}^\pi$ is affected by both the inherent transition probability $\mathcal{P}$ and the selected actions.

\textbf{Reward} $\mathcal{R}$: Reward $r_i^t$ is obtained after taking action $a_i^t$ at state $s_i^t$. The goal of DRL is to maximize the reward by learning an optimal action strategy $\pi$. Hence, designing an appropriate reward function $R$ based on objectives and constraints defined in Eq. (\ref{eq_first-objective})-(\ref{eq_constraints-6}) plays a significant role in solving MO-OPF via DRL. In this paper, a reward function $R(\cdot)$, composing of $8$ sub-reward functions from $r_1(\cdot)$ to $r_8(\cdot)$ is proposed to reduce power generation cost, alleviate voltage fluctuation, and satisfy operating constraints in power systems, which are formulated in Eq. (\ref{eqn_total_reward})-(\ref{eqn_reward_8}).
\begin{itemize}
    \item $r_1(\cdot)$ is defined as a negative reward function, representing power generation cost in correspondence with the first objective of MO-OPF in Eq. (\ref{eq_first-objective}), where $C_{\text{w},j}^{\text{d,r,p}}$ and $C_{\text{s},k}^{\text{d,r,p}}$ represent the summation of direct, reserve, and penalty cost of power generation for wind and solar PV power, respectively.
    \item Both $r_2(\cdot)$ and $r_3(\cdot)$ are designed to describe the power transmission loss, where $r^\text{l}$ is the standard line loss rate, and $I_i$ is the current value, together with its corresponding thermal limit $T_i$ \cite{10.5555/559961}.
    \item $r_4(\cdot)$ is proposed to control voltage fluctuation. $r_5(\cdot)$, $r_6(\cdot)$, and $r_7(\cdot)$ are developed based on MO-OPF constraints defined in Eq. (\ref{eq_constraints-1})-(\ref{eq_constraints-6}). Besides, $r_8(\cdot)$ aims to incentivize power systems to accommodate as many RERs as possible.
\end{itemize}
\begin{align}
    R &= \sum_{i=1}^8 w_ir_i, \quad r_i\gets \tilde{r}_i,\quad i=5,6,7, \label{eqn_total_reward}
    \\
    r_1 &= -\sum_{i=1}^{N_\text{t}} C_{\text{t},i}- \sum_{j=1}^{N_\text{w}} C_{\text{w},j}^{\text{d,r,p}}-\sum_{k=1}^{N_\text{s}} C_{\text{s},k}^{\text{d,r,p}},
    \\
    r_2 &= -\frac{1}{r^{\text{l}}}\times \frac{P^{\text{l}}}{\sum_{i=1}^{N^{\text{g}}}P_i},
    \\
    r_3 &= 1-\frac{1}{N_\text{b}}\sum_{i=1}^{N_\text{b}} \min\left(\frac{I_i}{T_i+\varepsilon},1\right),
    \\
    r_4 &= -F \label{eqn_voltage_reward},
    \\
    r_5 &= \frac{1}{N^{\text{g}}} \begin{cases}
    \sum_{i=1}^{N^{\text{g}}}\left(1-\frac{P_i}{\bar{P}_i}\right), \quad \forall{P_i}>\bar{P}_i,\\
    \sum_{i=1}^{N^{\text{g}}}\left(1-\frac{\munderbar{P}_i}{P_i}\right),\quad \forall{P_i}<\munderbar{P}_i,
    \end{cases}
    \\
    \tilde{r}_5 &= \exp(r_4)-1,
    \\
    r_6 &= \frac{1}{N^{\text{g}}} \begin{cases}
    \sum_{i=1}^{N^{\text{g}}}\left(1-\frac{Q_i}{\bar{Q}_i}\right), \quad \forall{Q_i}>\bar{Q}_i,\\
    \sum_{i=1}^{N^{\text{g}}}\left(1-\frac{\munderbar{Q}_i}{Q_i}\right),\quad \forall{Q_i}<\munderbar{Q}_i,
    \end{cases}
    \\
    \tilde{r}_6 &= \exp(r_5)-1,
    \\
    r_7 &= \frac{1}{N} \begin{cases}
    \sum_{i=1}^{N}(1-\frac{|\mathbb{V}_i|}{\bar{V}_i}), \quad \forall|\mathbb{V}_i|>\bar{V}_i,\\
    \sum_{i=1}^{N}(1-\frac{\munderbar{V}_i}{|\mathbb{V}_i|}),\quad \forall{|\mathbb{V}_i|}<\munderbar{V}_i,
    \end{cases}
    \\
    \tilde{r}_7 &= \exp(r_6)-1,
    \\
    r_8 &= \frac{\sum_{i=1}^{N_\text{w}+N_\text{s}}P_i}{\sum_{i=1}^{N_\text{w}+N_\text{s}}\bar{P}_i}. \label{eqn_reward_8}
\end{align}

\subsubsection{Solving MDP by DDPG}
\label{subsec_drl}
The objective of DRL is to maximize the expected reward $\bar{R}_\theta$ based on action policy $a=\pi_\theta(s)$, expressed as:
\begin{equation}
    \label{eq_drl-objective}
    \begin{aligned}
    \bar{R}_{\theta} &=\mathbb{E}_{a^t\sim\pi,r^t,s^{t+1}\sim \text{Env}}\left[R(\tau)\right]\\
    &= \sum_{\tau} R(\tau)P(\tau\mid \theta),
    \end{aligned}
\end{equation}
where $\theta$ represents the parameters of $\pi$, and $\tau$ is the trajectory of power flow reallocation, recording all the 4-tuple transitions $\{s^t,a^t,r^t,s^{t+1}\}$ from the beginning of $\tau$ to its end. 

We then introduce DDPG to maximize $\bar{R}_\theta$ \cite{ddpg}. DDPG is the most representative actor-critic DRL algorithm to solve MDP. The biggest difference between DDPG and other DRL algorithms is that the action policy $a=\pi_\theta(s)$ deterministically outputs the values of actions instead of the probability distribution of actions, which dramatically decreases the computation cost and makes it much easier to implement. Specifically, policy gradient method presented in Eq. (\ref{eq_policy-gradient}) is applied in DDPG to update our action policy.
\begin{gather}
    \label{eq_policy-gradient}
    \theta \gets \theta - \eta\nabla \bar{R}_\theta,\\
    \label{eq_reward-gradient}
    \nabla \bar{R}_\theta = \frac{1}{N_\tau} \sum_{n=1}^{N_\tau} \sum_{t=1}^{T_n} A^\theta\left(s^t_n,a^t_n\right)\nabla \log p\left(a^t_n|s^t_n,\theta\right),
\end{gather}
where $s^t_n$ and $a^t_n$ are state and action vector including all nodes' states and actions, respectively formulated as
\begin{gather}
    s^t_n = \textbf{concat}\left[\left(s^t_1, s^t_2, \cdots, s^t_N, \bm{y}^t\right)\right],\\
    a^t_n = \textbf{concat}\left[\left(a^t_1,a^t_2,\cdots,a^t_N\right)\right],
\end{gather} 
and note that $s^t_n$ also contains the result of preceding MG-ASTGCN $\bm{y}^t$ defined in Eq. (\ref{eq_output-mgastgcn}). 

Advantage function $A^\theta\left(s^t_n,a^t_n\right)$ is a criterion to assess how good the chose action $a^t_n$ is under current state $s^t_n$ compared to baseline, which can be formulated as
\begin{gather}
\label{eqn_adv_func}
    \mathcal{A}^\theta\left(s^t_n,a^t_n\right) =  G^t_n -b,\\
    G^t_n = \sum_{t'=t}^{T_n}\gamma^{t'-t}r_{t'}^n,\quad b = \mathbb{E}_{s\sim \text{Env}}\left[R(s)\right],
\end{gather}
where $\gamma$ is a discounting factor, and $b$ represents the baseline of reward considering all possible actions. Due to uncertainties of DRL environments, both $G^t_n$ and $b$ are obviously random variables. Guaranteeing accurate estimation of advantage function, DDPG has proposed a critic network $Q^\pi (a^t,a^t)$ which can be formulated in Eq. (\ref{eq_critic-net}). Hence, the advantage function can be rewritten in Eq. (\ref{eq_rewrite-adv-func}).
\begin{gather}
\label{eq_critic-net}
Q^\pi (s^t,a^t) = \mathbb{E}_{r^t,s^{t+1}\sim \text{Env}}\left\{r^t+\gamma Q^\pi\left[s^{t+1},\pi_\theta\left(s^{t+1}\right)\right]\right\},\\
\label{eq_rewrite-adv-func}
A^\theta\left(s^t_n,a^t_n\right) = Q^\pi\left(s^t_n,a^t_n\right)-\sum_{a^t_n}Q^\pi\left(s^t_n,a^t_n\right).
\end{gather}

\begin{figure}[!b]
    \centering
    \vspace*{-6mm}
    \includegraphics[width=3.2in,height=1.7in]{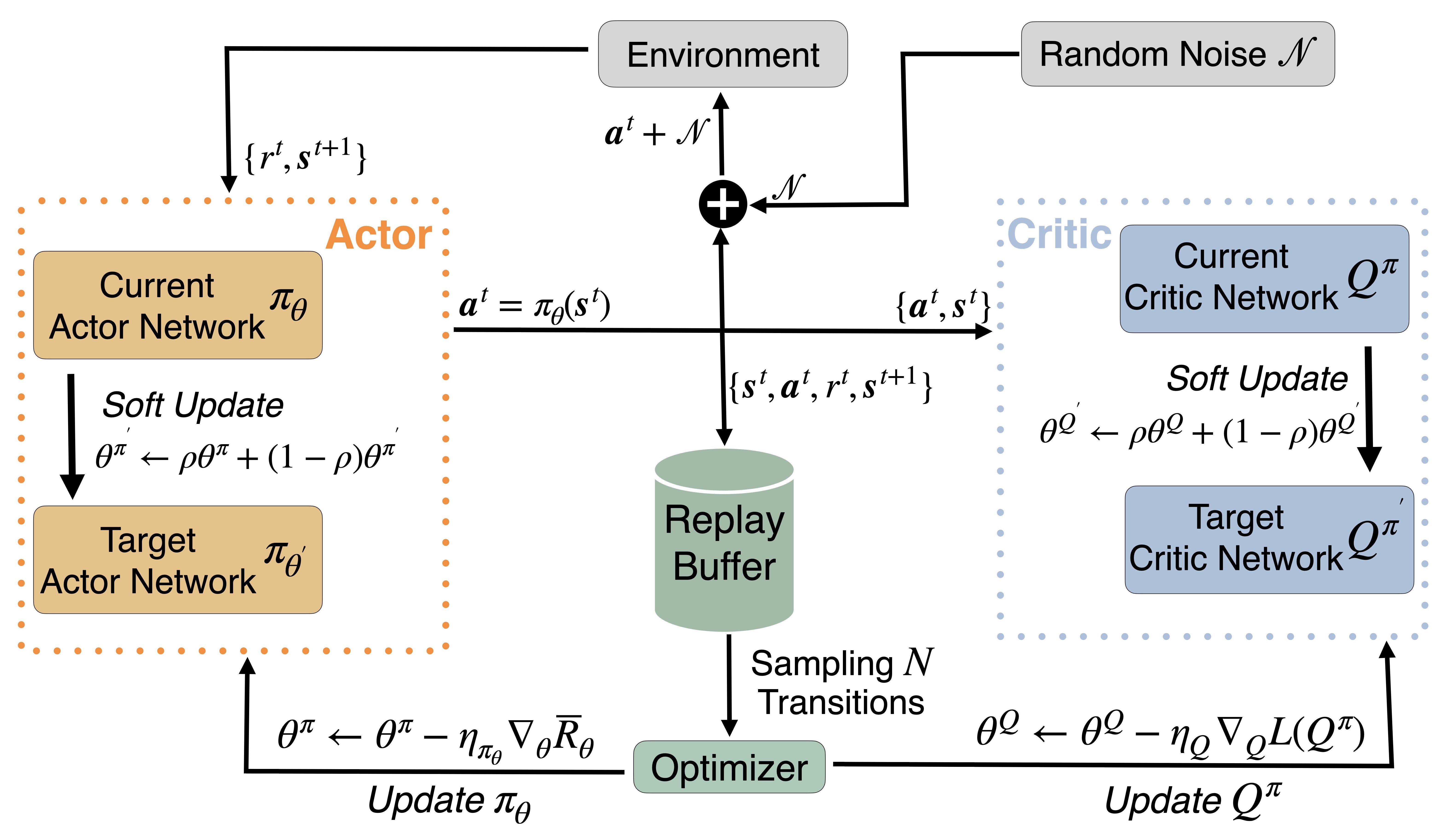}
    \vspace*{-4mm}
    \caption{The workflow of DDPG.}
    \label{fig:ddpg_workflow}
\end{figure}

Since calculating $Q^\pi$ depends on environments rather than our action policy $\pi$, it is applicable to learn $Q^\pi$ in an off-policy way of using transitions generated from a different action policy $\pi_{\theta'}$. The critic network can be updated by minimizing the root mean square error formulated in Eq. (\ref{eq_critic-net-loss-func}). The workflow of DDPG is presented in Fig. \ref{fig:ddpg_workflow}.
\begin{gather}
\label{eq_critic-net-loss-func}
L\left(Q^\pi\right) = \mathbb{E}_{s^t,r^t\sim \text{Env}, a^t\sim \pi_{\theta'}} \left\{\left[Q^\pi\left(s^t,a^t\right)-q^t\right]^2\right\},\\
q^t = r^t + \gamma Q^{\pi'}\left[s^{t+1},\pi_{\theta'}\left(s^{t+1}\right)\right].
\end{gather}

\textbf{Termination Criterion of DDPG's Training:} Solving the OPF is based on the fact that power flow has reached convergence. Therefore, at each timestep $t$ during DDPG's training process, training should be terminated and restrated if the power flow is not converged. The convergence condition \cite{10.5555/559961} is formulated as
\begin{equation}
    \lVert \bm{v}_i^t-\bm{v}_i^{t-1} \rVert \leq \epsilon,\quad \forall i\in \mathcal{V},
\end{equation}
where $\epsilon$ is the mismatch tolerance to measure the distance between $\bm{v}_i^t$ and $\bm{v}_i^{t-1}$. $\bm{v}_i^t$ and $\bm{v}_i^{t-1}$ represent characteristics of the $i$th node in $\mathcal{G}$ at power flow reallocation interval $t$ and $t-1$, respectively, formulated as
\begin{align}
    \bm{v}_i^{t}&=\left(P_i^t,Q_i^t,\mathbb{V}_i^t\right)^T,\\
    \bm{v}_i^{t-1}&=\left(P_i^{t-1},Q_i^{t-1},\mathbb{V}_i^{t-1}\right)^T.
\end{align}

Additionally, early stopping mechanism is applied when implementing DDPG, calculating the cumulative reward $r_{\text{c}}$ for every training episode $n$, in which $r_c$ can be formulated as
\begin{equation}
    r_{\text{c}} = \sum_{t=1}^{t'} r^t,
\end{equation}
where $t'$ represents the current training timestep.

Once the $r_{\text{c}}$ exceeds the early stopping's threshold $r_{\text{es}}$, its corresponding training episode is terminated. The early stopping mechanism aims to not only prevent DDPG from learning unsatisfactory policy to reallocate the power flow, but also accelerate its convergence speed. 

In summary, DDPG follows the policy-gradient way to solve the derived MDP, where a critic network is introduced to assess our learned agent action policy. The detailed algorithmic procedure of our DRL-based strategy for solving the MO-OPF problem is presented in Algorithm \ref{alg_ddpg}.

\begin{algorithm}
\caption{The DRL-based Strategy for Solving the MO-OPF Problem} 
\label{alg_ddpg}
\begin{algorithmic}
\STATE Randomly initialize action policy $\pi_\theta(s)$ and critic network $Q^\pi(s,s)$ with weights $\theta^Q$ and $\theta^\pi$
\STATE Initialize target networks $\pi_{\theta'}$ and $Q^{\pi'}$ with weights $\theta^{Q'}\gets \theta^Q$, $\theta^{\pi'}\gets \theta^{\pi}$
\STATE Initialize the replay buffer $\mathcal{B}$ with sufficient capacity
\FOR{$n=1,2,\cdots,N_\tau$}
\STATE Initialize a random process $\mathcal{N}$ for action exploration
\STATE Reset the original state $s^1$
\STATE Reset the cumulative reward $R_\text{c}\gets 0$
\FOR{$t=1,2,\cdots,T_n$}
\STATE Receive the output of MG-ASTGCN $\bm{y}^t$ and concatenate it with $s^t$
\STATE Select action $s^t$ = $\pi_\theta(s^t)$ + $\mathcal{N}$ based on current action policy $\pi^\theta$ and exploration noise $\mathcal{N}$
\STATE Interact with the environment using action $s^t$ and get reward $r^t$ and the next state $s^{t+1}$
\IF{$\lVert \bm{v}_i^t-\bm{v}_i^{t-1} \rVert > \epsilon$}
\STATE \textbf{Break}
\ENDIF
\STATE Update the cumulative reward $r_{\text{c}} \gets r_{\text{c}}+r^t$
\IF{$r_{\text{c}}>r_{\text{es}}$}
\STATE \textbf{Break}
\ENDIF
\STATE Store a transition $\{s^t,s^t,r^t,s^{t+1}\}$ in $\mathcal{B}$
\STATE Randomly sample a batch of $N$ transitions from $\mathcal{B}$.
\STATE Calculate $q^t$ using the target networks, $q^t\gets r^t+\gamma Q^{\pi'}[s^{t+1},\pi_{\theta'}(s^{t+1})]$
\STATE Update the critic network $Q^\pi$ by using gradient descent to minimizing the RMSE function $L(Q^\pi) = \frac{1}{N}\sum_i [Q^\pi(s^t,s^t)-q^t]^2$:
\STATE $\qquad \qquad \qquad \theta^Q \gets \theta^Q-\eta_Q\nabla_QL(Q^\pi)$
\STATE Update the action policy $\pi_\theta$ by policy gradient:
\STATE $\qquad \qquad \qquad \theta^\pi\gets \theta^\pi-\eta_{\pi_\theta}\nabla_\theta\bar{R}_\theta$
\STATE Update target networks using the soft update method with its corresponding parameter $\rho$:
\STATE $\qquad \qquad \qquad \theta^{Q'}\gets \rho\theta^Q+(1-\rho)\theta^{Q'}$
\STATE $\qquad \qquad \qquad \theta^{\pi'}\gets\rho\theta^{\pi}+(1-\rho)\theta^{\pi'}$.
\ENDFOR
\ENDFOR
\end{algorithmic}
\end{algorithm}
\begin{figure}
    \centering
    \includegraphics[width=3.4in,height=2.1in]{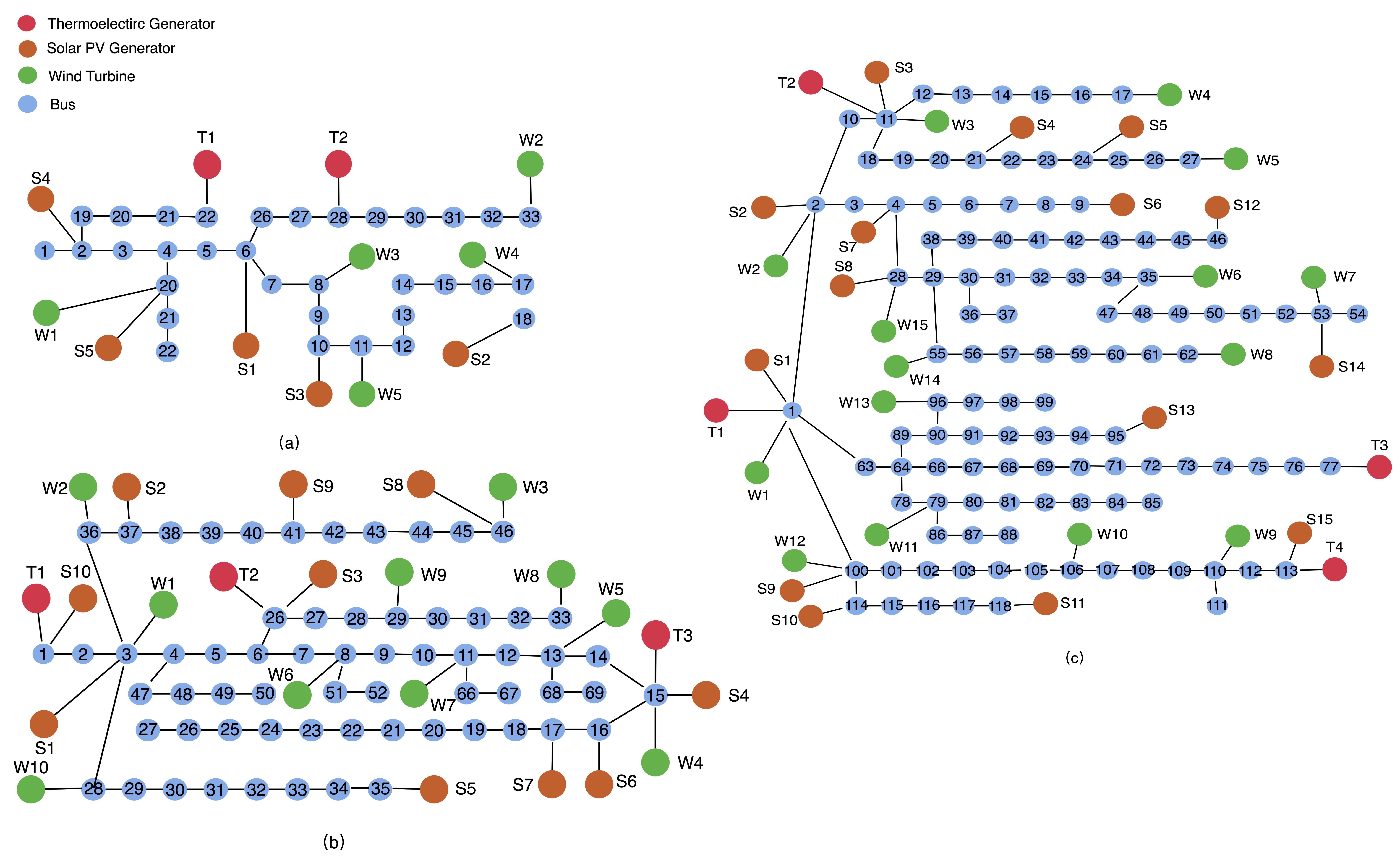}
    \vspace*{-4mm}
    \caption{The structures of modified IEEE 33-bus, 69-bus, and 118-bus RDSs, shown in (a), (b), and (c), respectively.}
    \label{fig:ieee_model}
    \vspace*{-7mm}
\end{figure}
\section{Experiments and Results}
\label{sec_experiments}
\subsection{Experimental Settings}
\label{subsec_experimental-settings}
\subsubsection{Application Scenario}
The proposed DRL-based strategy is tested on the modified IEEE $33$-bus, $69$-bus, and $118$-bus radial distribution systems (RDSs), as illustrated in Fig. \ref{fig:ieee_model}. The three RDSs characteristics consist of the number of generators, baseline voltage, baesline apparent power, load active power, and load reactive power, as presented in Table \ref{tab:ieee_model}.Besides, a workstation with $5$ Nvidia TITAN RTX graphics processing units is used for the DRL training.
\begin{table}[!ht]
    \centering
    \vspace*{-4mm}
    \caption{System Charateristics in IEEE $33$, $69$, and $118$-bus RDSs.}
    \vspace*{-2mm}
    \begin{tabular}{ M{4.2cm}  M {0.9cm}  M{0.9cm}  M{1.1cm} }
    \hline
    \textbf{RDSs Charateristics}  &  $\bm{33}$\textbf{-Bus}    & $\bm{69}$\textbf{-Bus}  & $\bm{118}$\textbf{-Bus}\\
    \hline
    Total Buses ($N_l$)  &   $33$  &   $69$  &   $118$\\[0.5ex]
    Thermoelectric Generators ($N_t$)   &   $2$   &   $3$   &$4$\\[0.5ex]
    Wind Turbines ($N_w$) &   $5$   &   $10$  &   $15$\\[0.5ex]
    Solar PV Generators ($N_s$) &   $5$   &   $10$  &   $15$\\[0.5ex]
    Baseline Voltage (kV)   &   $12.66$ &   $12.66$ &   $11$\\[0.5ex]
    Baseline Apparent Power (MVA)   &   $100$   &   $100$   &   $100$\\[0.5ex]
    Total Load Active Power (MW)    &   $3.715$  &   $3.800 $  &   $22.710$\\[0.5ex]
    Total Load Reactive Power (MVAR)    &   $2.300$   &   $2.690$   &   $17.041$\\[0.5ex]
    \hline
    \end{tabular}
    \label{tab:ieee_model}
    \vspace*{-2mm}
\end{table}

\subsubsection{Algorithm Performance Metric}
In our experiments, reward functions defined in Eq. (\ref{eqn_total_reward})-(\ref{eqn_reward_8}) are adopted to measure the performance of both DDPG and benchmark algorithms. Specifically, we introduce a criterion to assess algorithms for solving the MO-OPF problem, and the criterion is formulated as
\begin{equation}
\label{eq_score}
    \text{SCORE} = \frac{1}{N_{\text{eval}}}\sum_{n=1}^{N_{\text{eval}}}\sum_{t=1}^{T_{\text{end}}} R^t_n,
\end{equation}
where $N_{\text{eval}}$ is the number of episodes for evaluation, and $T_{\text{end}}$ is the length of each episode. Both $N_{\text{eval}}$ and $T_{\text{end}}$ are initialized as $100$. The detailed process of evaluation is presented in Algorithm \ref{alg_test}.
\begin{table*}[!t]
    \centering
    \caption{The average testing time and evaluation results of the proposed DRL-based strategy, HHO, and GWO algorithms.}
    \vspace*{-2mm}
    \begin{tabular}{M{2cm} M{2cm} M{2cm} ||M{2cm} M{2cm}|| M{2.2cm} M{2cm}}
    \hline
    Model     & Average Testing Time in 33-bus RDS &SCORE in 33-bus RDS & Average Testing Time in 69-bus RDS& SCORE in 69-bus RDS& Average Testing Time in 118-bus RDS&SCORE in 118-bus RDS\\
    \hline
    HHO  & $2.35s$ /timestep & $3530.66$& $6.12s$ /timestep& $6104.81$ & $12.69s$ /timestep & $8231.95$ \\[0.5ex]
    GWO  & $3.44s$ /timestep & $3014.44$ & $8.12s$ /timestep & $5821.99$ & $15.79s$ /timestep  & $7481.24$\\[0.5ex]
    \textbf{DRL (Ours)}   & $\bm{0.89s}$ /timestep & $\bm{4014.59}$ & $\bm{1.26s}$ /timestep & $\bm{7756.64}$ & $\bm{1.57s}$ /timestep & $\bm{14384.11}$ \\[0.5ex]
    \hline
    \end{tabular}
    \label{tab:heuristic_results}
    \vspace*{-7mm}
\end{table*}
\begin{figure*}[!t]
\centering
\subfloat[IEEE $33$-bus RDS]{
\includegraphics[width=2.2in,height=1.45in]{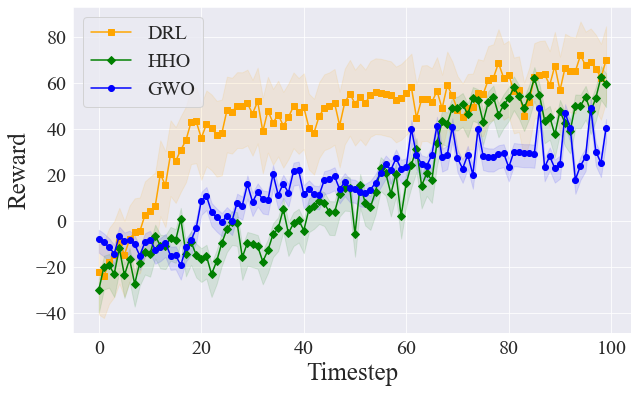}
\label{fig:heuristic-33}}
\subfloat[IEEE $69$-bus RDS]{
\includegraphics[width=2.2in,height=1.45in]{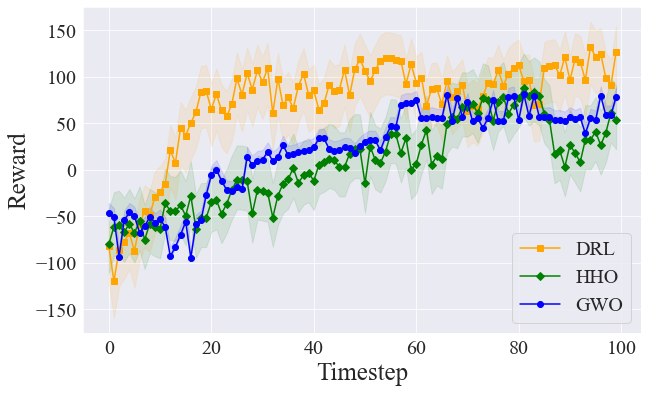}
\label{fig:heuristic-69}}
\subfloat[IEEE $118$-bus RDS]{
\includegraphics[width=2.2in,height=1.45in]{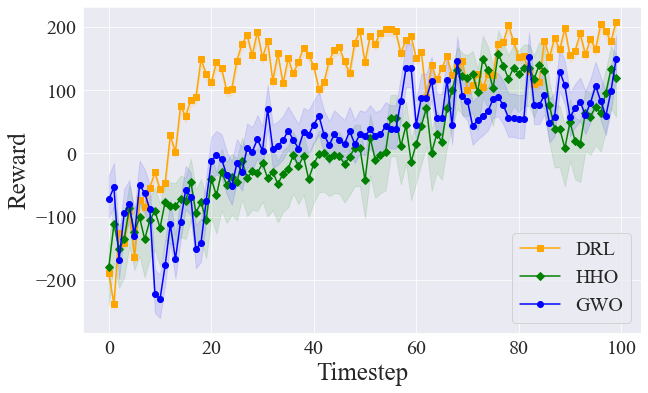}
\label{fig:heuristic-118}}
\vspace*{-2mm}
\caption{Evaluation results of HHO, GWO, and DRL-based strategy on three testing RDSs.}
\label{fig:heuristic_results}
\vspace*{-7mm}
\end{figure*}
\begin{figure*}[!t]
\centering
\subfloat[IEEE $33$-bus RDS]{
\includegraphics[width=2.2in,height=1.45in]{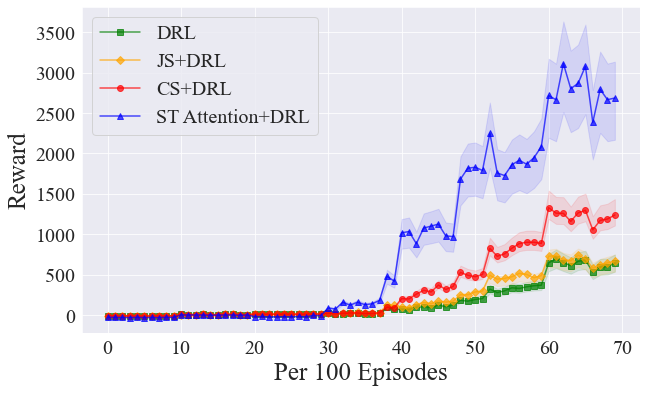}
\label{fig:att-33}}
\subfloat[IEEE $69$-bus RDS]{
\includegraphics[width=2.2in,height=1.45in]{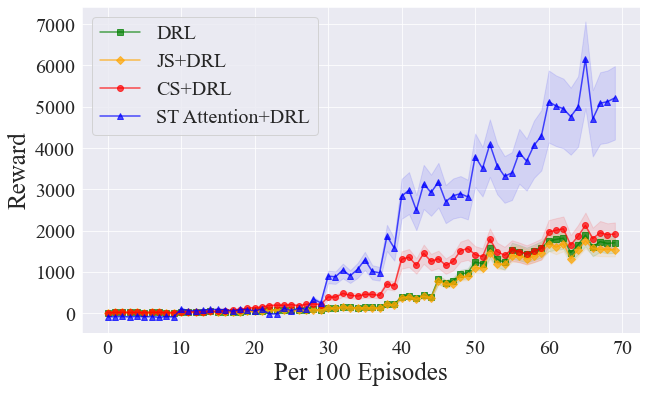}
\label{fig:att-69}}
\subfloat[IEEE $118$-bus RDS]{
\includegraphics[width=2.2in,height=1.45in]{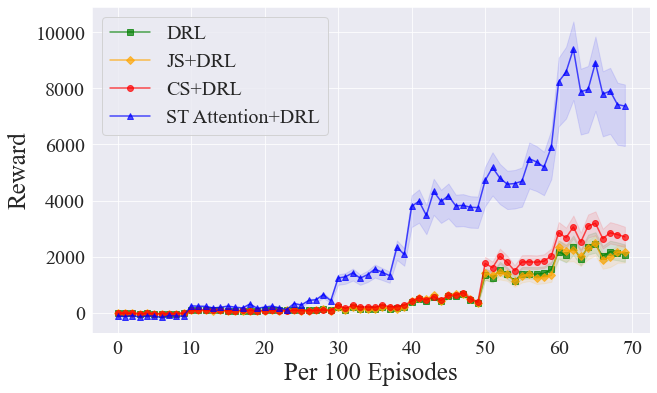}
\label{fig:att-118}}
\vspace*{-2mm}
\caption{Evaluation results of HHO, GWO, and DRL-based strategy on three testing RDSs.}
\label{fig:att_results}
\vspace*{-5mm}
\end{figure*}

\subsection{Experimental Results}
\label{subsec_experimental-results}
\subsubsection{Comparisons}
Two representative heuristic algorithms---harris hawk optimization (HHO)\cite{hho} and grey wolf optimization (GWO)\cite{gwo}, are adopted to solve the MO-OPF problem. The evaluation results of these two heuristic algorithms, together with the proposed DRL-based strategy on IEEE $33$, $69$, and $118$-bus RDSs are illustrated in Fig. \ref{fig:heuristic_results} and Table \ref{tab:heuristic_results}. We see that the proposed DRL-based strategy outperforms other two heuristic algorithms. We also find that DRL-based strategy consumes much less time at each timestep while still achieving outstanding performance. The reason for such performance gap is twofold:
\begin{itemize}
    \item Heuristic algorithms do not need training, resulting in longer computation time at each timestep. Meanwhile, since they tend to get stuck in local optimums, their selected actions are less likely to be the optimal ones, resulting in smaller rewards.
    \item Tremendous data sampled by the DRL-based strategy results in its more effective and efficient searching in the action space. Therefore, DRL reacts much faster at the beginning of evaluation and gradually obtains a higher reward, especially in the large-scale power system, as shown in Fig. \ref{fig:heuristic-118}.
\end{itemize}

\subsubsection{Effectiveness of ST Attention}
To evaluate the effectiveness of the ST attention and its impact on the sequential DDPG, the ST attention mechanism is substituted with several other techniques for graphical correlation extraction, including cosine similarity (CS) and jaccard similarity (JS), whose training results are presented in Fig. \ref{fig:att-118}. We can summarize several observations regarding the effectiveness of the ST attention:
\begin{itemize}
    \item The adoption of the ST attention can dramatically increase DDPG's convergence speed, where more effective searching in the action space is conducted based on the extracted graphical knowledge
    \item It is challenging for a standard DRL algorithm to tackle the complex MO-OPF problem, since both action and state space in RDSs are considerably large.
    \item The substituted methods are less effective than the ST attention, since they only focus on degree correlations among different nodes, ignoring both nodes' inner features and temporal dependencies.
\end{itemize}

\begin{figure}[!t]
    \centering
    \includegraphics[width=3.4in,height=1.47in]{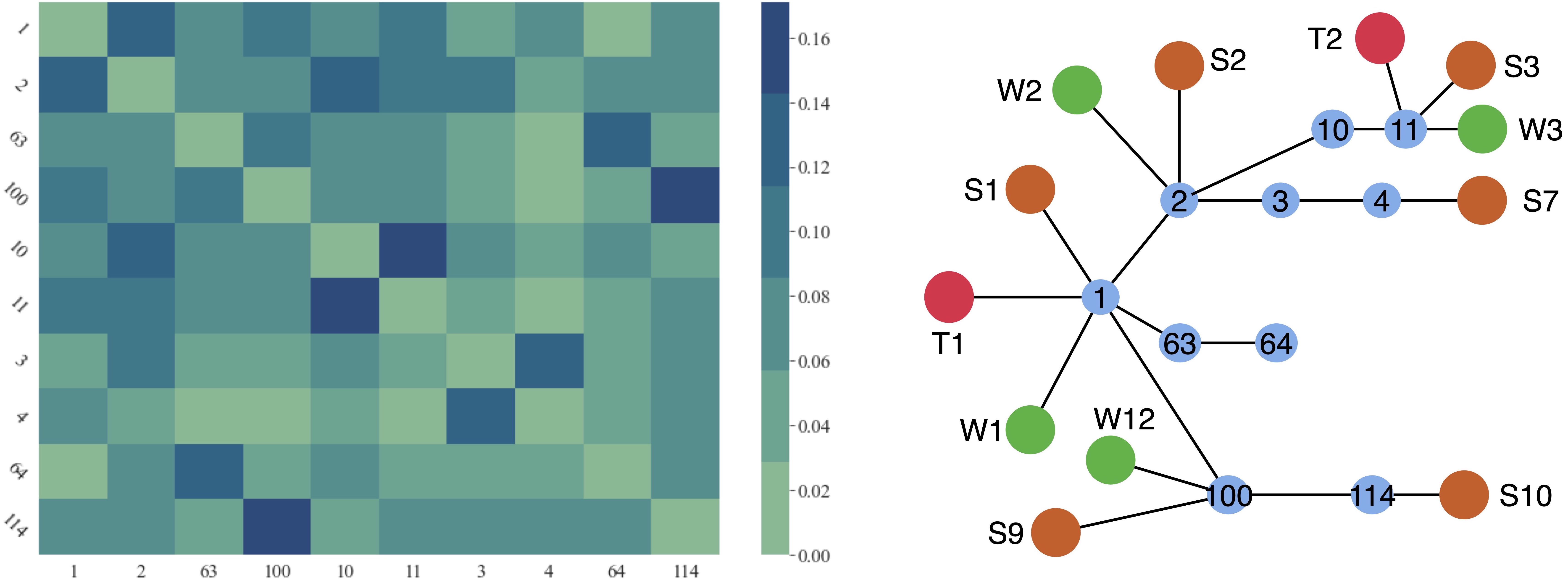}
    \vspace*{-3mm}
    \caption{The partial spatial attention matrix and its corresponding sub-graph in IEEE $118$-bus RDS.}
    \label{fig:satt_mat}
    \vspace*{-4mm}
\end{figure}

\begin{figure}[!t]
    \centering
    \includegraphics[width=3.4in,height=1.7in]{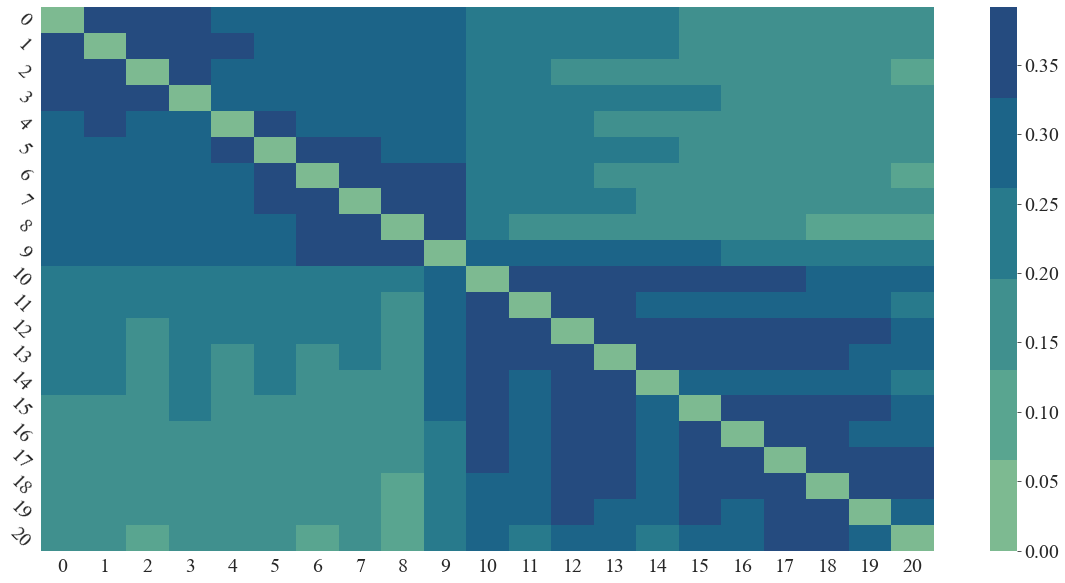}
    \vspace*{-3mm}
    \caption{The partial temporal attention matrix of Bus $1$'s recent segment in IEEE 118-bus RDS.}
    \label{fig:tatt_mat}
    \vspace*{-5mm}
\end{figure}

Additionally, we observe two interesting phenomena from spatial and temporal attention matrices, which are illustrated in Fig. \ref{fig:satt_mat} and Fig. \ref{fig:tatt_mat}, respectively.
\begin{itemize}
    \item In Fig. \ref{fig:satt_mat}, the spatial attention mechanism tends to focus on node pairs with more generator access, which means larger weights are assigned to their correlation strengths. For instance, although Bus $2$ and $114$ are not adjacency, they can be connected by Bus $1$ and $100$, where $2$ and $3$ power generators are connected, respectively. Therefore, it is reasonable that the correlation between Bus $2$ and $114$ is more significant than other nonadjacent node pairs.
    \item In Fig. \ref{fig:tatt_mat}, in the recent temporal attention, the correlation strengths between current and previous node features drop sharply when it comes to the $10$th previous node feature vector, where we can conclude that the latest $10$ feature vectors of one node contain more significant temporal correlation information.
\end{itemize}

\subsubsection{Stability Test of DRL-based strategy} 
\begin{figure}[!ht]
\vspace*{-4mm}
\centering
\subfloat[]{
\includegraphics[width=1.65in,height=1.3in]{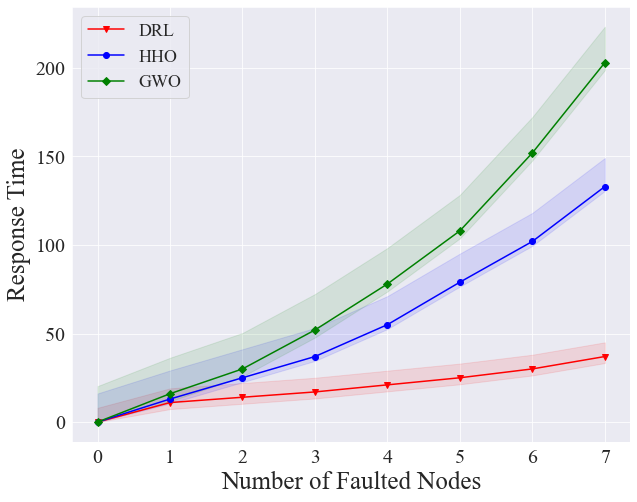}
\label{fig:different-response}}
\subfloat[]{
\includegraphics[width=1.65in,height=1.3in]{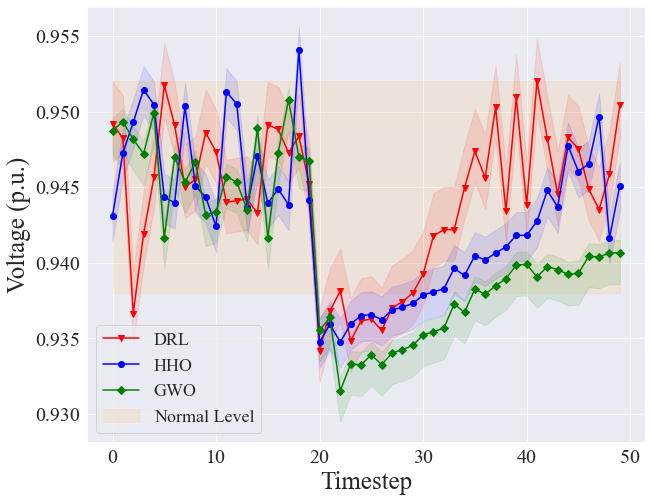}
\label{fig:specific-response}}
\vspace*{-2mm}
\caption{(a) Response time when facing different number of faulted nodes in IEEE 69-bus RDS; (b) Case study of node fault: one node failed in IEEE 69-bus RDS.}
\label{fig:stability}
\end{figure}

The average response time is defined to assess the stability of power systems when facing node faults, which can be calculated by counting how many timesteps that the power system takes to recover voltage to its normal level. Fig. \ref{fig:different-response} illustrates the response time of HHO, GWO, and the proposed DRL-based strategy with different numbers of faulted nodes. We see in Fig. \ref{fig:different-response} that the response time of the DRL algorithm grows linearly compared to those of other two heuristic algorithms increasing exponentially. Besides, Fig. \ref{fig:specific-response} shows a more detailed case study with one faulted node in the IEEE $69$-bus RDS, indicating that power systems' stability can be significantly improved using our DRL-based strategy.

\subsubsection{Voltage Fluctuation Control}
\begin{figure}[!t]
    \centering
    \includegraphics[width=3in,height=2.4in]{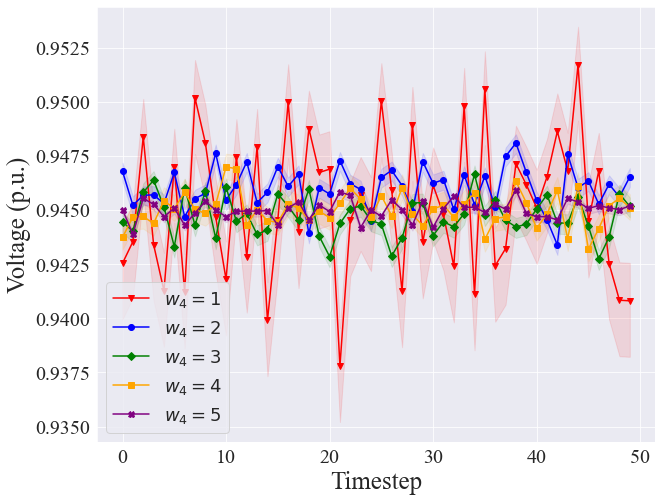}
    \vspace*{-2mm}
    \caption{Voltage fluctuation of Bus $15$ in IEEE $69$-bus RDS with different voltage fluctuation weight $w_4$ defined in Eq. \ref{eqn_voltage_reward}.}
    \label{fig:voltage-fluctuation}
    \vspace*{-5mm}
\end{figure}

The weights assigned for each reward function described in Eq. (\ref{eqn_total_reward})-(\ref{eqn_reward_8}) represent their corresponding importance. We conduct several experiments with different \emph{voltage fluctuation weight} $w_4$, to test the DRL-based strategy's capability in dealing with voltage fluctuations, as presented in Table \ref{tab:weights}. Specifically, Bus $15$'s voltage fluctuation in IEEE $69$-bus RDS is presented in Fig. \ref{fig:voltage-fluctuation}. Interestingly, although the voltage fluctuation is well-controlled with the increase of $w_4$, as shown in Fig. \ref{fig:voltage-fluctuation}, our DRL-based strategy seems to get stuck in local optima, i.e., SCOREs with $w_4$ from $2$ to $4$ are lower than $w_4$ initialized as $1$. Moreover, we find that the performance of DRL-based strategy degenerates with larger $w_4$, when it comes to large-scale power systems. The proposed DRL-based strategy will find the sub-optimal power flow, if voltage fluctuation control is overemphasized.
\begin{table}[!t]
    \centering
    \vspace*{-2mm}
    \caption{The evaluation results with different voltage fluctuation weight.}
    \vspace*{-2mm}
    \begin{tabular}{M{1cm} M{2cm} M{2cm} M{2cm}}
    \hline
    $w_4$ &SCORE in 33-bus RDS & SCORE in 69-bus RDS&SCORE in 118-bus RDS\\
    \hline
    $\bm{1}$  & $\bm{4014.59}$ & $\bm{7756.64}$ & $\bm{14384.11}$\\[0.5ex]
    2  &  $3979.14$&  $7544.81$ & $13958.24$ \\[0.5ex]
    3  &  $3967.51$ &  $7394.53$ &  $13572.31$\\[0.5ex]
    4  & $3843.72$ &  $7252.66$ &  $12891.59$ \\[0.5ex]
    5  & $3687.13$ &  $7036.29$ &  $12186.67$ \\[0.5ex]
    \hline
    \end{tabular}
    \label{tab:weights}
\end{table}

Interestingly, although voltage fluctuation is well-controlled with the increase of $w_4$ as shown in Fig. \ref{fig:voltage-fluctuation}, our DRL-based strategy seems to get stuck in local optimums, i.e., SCOREs with $w_4$ from $2$ to $4$ are lower than $w_4$ initialized as $1$. Moreover, based on our experiments, we find that the performance of DRL-based strategy degenerates with larger $w_4$ when it comes to large-scale power systems. We can conclude that the proposed DRL-based strategy will find the sub-optimal power flow if voltage fluctuation control is overemphasized.
\begin{algorithm}
\caption{Calculating SCORE for algorithm evaluation} 
\label{alg_test}
\begin{algorithmic}
\STATE Initialize $\text{SCORE} \gets 0$\\
\FOR{$n=1,2,\cdots,N_{\text{eva}}$}
\STATE Reset original state $s^1$\\
\STATE Initialize cumulative reward $R_n\gets 0$\\
\FOR{$t=1,2,\cdots,T_{\text{end}}$}
\STATE Determine current action $a^t$ based on the input algorithm\\
\STATE Get reward $r^t$ and $s^{t+1}$ by interacting with $\mathcal{G}$\\
\STATE Update cumulative reward $R_n \gets R_n+r^t$\\
\IF{$\lVert \bm{v}_i^t-\bm{v}_i^{t-1} \rVert > \epsilon$}
\STATE \textbf{Break}
\ENDIF
\ENDFOR
\STATE Update SCORE: \\
\STATE $\qquad \qquad \qquad \text{SCORE}\gets \text{SCORE}+R_n$
\ENDFOR
\STATE Calculate average SCORE:
\STATE $\qquad \qquad \qquad \text{SCORE}\gets \frac{1}{N_{\text{eva}}}\text{SCORE}$.
\end{algorithmic}
\end{algorithm}

\section{Conclusions and Future Works}
\label{sec_conclusions}
In this paper, we proposed a DRL-based strategy, accompanied by multi-grained ST graph information. The aim is to alleviate uncertainties brought in by RERs and to improve power systems' stability for solving the OPF problem more effectively and efficiently. First, we derive the MOO-OPF formulation considering the high renewable penetration in power systems. Then, to make full use of the ST features and correlations in DNs, MG-ASTGCN is proposed to extract ST information in multi-time scales. In the end, we adopt DDPG to solve the complex MOO-OPF problem. We can draw several conclusion based on experimental results: (i) Extracting ST correlations in power systems plays an essential role in solving the MOO-OPF problem, where the performance of the DRL-based strategy degenerate significantly without ST attention mechanism; (ii) Compared with several heuristic algorithms, the proposed DRL-based strategy achieves better performance in solving the OPF problem with less computational time. Besides, the adoption of the DRL-based strategy improves power systems' stability, which has a shorter response time when facing node faults; (iii) In power systems, node pairs with more generator access seem to have a stronger spatial correlation. Moreover, in the temporal view, we can conclude that the latest 10 feature vectors of nodes contain more valuable temporal correlation information; (iv) Finding OPF and dealing with technical problems seems to be a trade-off, since our experimental results indicate that overemphasizing voltage fluctuation control results in sub-optimal operations.

In our future work, designing a suitable incentive mechanism to accommodate more RERs in smart grids will be studied.

\bibliographystyle{IEEEtran}
\bibliography{IEEEabrv}
\end{document}